\def\year{2022}\relax
\pdfoutput=1
\documentclass[letterpaper]{article} 
\usepackage{aaai22}  
\usepackage{times}  
\usepackage{helvet}  
\usepackage{courier}  
\usepackage[hyphens]{url}  
\usepackage{graphicx} 
\usepackage{natbib}  
\usepackage{caption} 
\DeclareCaptionStyle{ruled}{labelfont=normalfont,labelsep=colon,strut=off} 
\usepackage{algorithm}
\usepackage{algorithmic}
\usepackage{booktabs}
\usepackage{multirow}
\usepackage{amsmath}
\usepackage{subfig}
\usepackage{newfloat}
\usepackage{listings}
\floatstyle{ruled}
\newfloat{listing}{tb}{lst}{}
\floatname{listing}{Listing}
\title{Interpretable Low-Resource Legal Decision Making}
\author {
    Rohan Bhambhoria\textsuperscript{\rm 1,}\textsuperscript{\rm 2,}\textsuperscript{\rm 3},
    Hui Liu \textsuperscript{\rm 1,}\textsuperscript{\rm 2},
    Samuel Dahan \textsuperscript{\rm 1,}\textsuperscript{\rm 3,}\textsuperscript{\rm 4},
    Xiaodan Zhu \textsuperscript{\rm 1,}\textsuperscript{\rm 2,}\textsuperscript{\rm 3}
}
\affiliations {
    \textsuperscript{\rm 1} Ingenuity Labs \& \textsuperscript{\rm 2} ECE , Queen's University\\
    \textsuperscript{\rm 3} Conflict Analytics Lab, Queen's University\\
    \textsuperscript{\rm 4} Cornell Law School \\
    \{r.bhambhoria, hui.liu, samuel.dahan, xiaodan.zhu\}@queensu.ca
}

\begin{document}

\maketitle


\begin{abstract}
Over the past several years, legal applications of deep learning have been on the rise. However, as with other high-stakes decision making areas, the requirement for interpretability is of crucial importance. Current models utilized by legal practitioners are more of the conventional machine learning type, wherein they are inherently interpretable, yet unable to harness the performance capabilities of data-driven deep learning models. In this work, we utilize deep learning models in the area of trademark law to shed light on the issue of likelihood of confusion between trademarks. Specifically, we introduce a model-agnostic interpretable intermediate layer, a technique which proves to be effective for legal documents. Furthermore, we utilize weakly supervised learning by means of a curriculum learning strategy, effectively demonstrating the improved performance of a deep learning model. This is in contrast to the conventional models which are only able to utilize the limited number of expensive manually-annotated samples by legal experts. Although the methods presented in this work tackles the task of risk of confusion for trademarks, it is straightforward to extend them to other fields of law, or more generally, to other similar high-stakes application scenarios. Our code is available here\footnote{https://github.com/RohanVB/Interpretable-Low-Resource-Legal-Decision-Making}.
\end{abstract}

\section{Introduction}

In many legal tasks, a thorough review of case law is required to understand the underlying elements involved in the legal decision-making process. The use-case of an automatic system capable of addressing the aforementioned varies, and ranges from promoting access to justice for self-represented litigants to determining precedents which may act favourably for legal professionals \cite{10.2307/25722338}.

\begin{figure}[t]
\centering
\includegraphics[width=0.85\columnwidth]{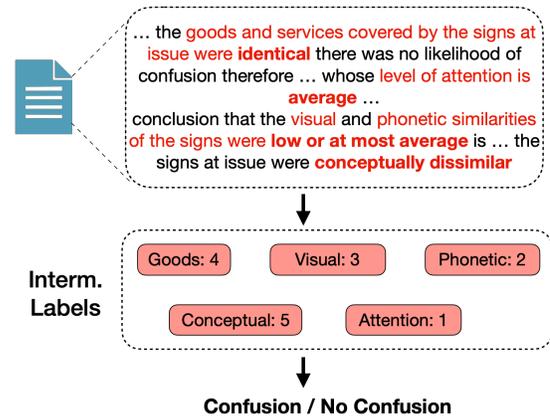} 
\caption{For a given case law document, important factors involved are{\textemdash}the similarity of goods \& services, phonetic similarity, conceptual similarity, visual similarity, and the level of attention of the relevant public. These factors are sufficient to decide the outcome of a case, and we denote them as ``intermediate labels" in this work.}
\label{fig1}
\end{figure}

In this work, for purposes of exploring a real-world scenario, we utilize cases from the likelihood of confusion for trademarks, a branch of intellectual property law. Likelihood of confusion exists between trademarks when the marks are so similar and the goods and/or services for which they are used are so related that consumers would mistakenly believe they come from the same source \cite{Bone2012TakingTC}. Accordingly, this task involves analyzing the goods and services provided by each company in order to identify the overlap. In addition, the marks of the companies are analyzed from several different perspectives{\textemdash}how they differ visually, phonetically, and conceptually to influence the judgment as shown in Figure \ref{fig1}. Finally, other less important factors such as the level of attention to the relevant public are considered. The case law hence provides a detailed discussion along with reasons and conclusions which favor the judgment process \cite{samdahan2020}. 

Assessing the likelihood of confusion for trademark is a crucial component of a company's existence, especially for IP-intensive small to medium sized companies (SMEs) that need to protect their creation. Firms with registered intellectual property (IP) rights perform better than those without,  yet few SMEs in the EU actually exercise their IP rights: only 9\% of SMEs have registered IP rights, compared to 36\% of large companies. However, research suggests that SMEs generally do not use the IP system because they do not see the benefit of it, they lack the necessary expertise, and/or they find the procedure too costly \cite{samdahan2020}. As a result, many of these firms often incur unforeseen costs associated with litigation to defend their mark, either before or after it is registered. 

Considering that innovative SMEs are a primary driver of growth, it is worth exploring new ways to democratize IP legal help. As a result, research in this space aims to contribute to this endeavour, and recent work \cite{samdahan2020,samdahan2022} discusses the potential implications of this field of research in a recent expert opinion for the European Union Intellectual Property Office, the higher regulatory authority in Europe for IP regulation. In particular, in consultation with regulators and practitioners, they look at a new generation of direct-to-public (DTP) AI tools that can improve access to justice by providing the minimal level of legal help necessary for meeting basic legal needs. The authors review how a combination of online dispute resolution (ODR) and predictive analytics employing machine learning and deep learning can positively affect EU trademark dispute resolution. The system operates as a legal aid tool for small to medium-sized enterprises (SMEs) to determine potential IP rights violations. 

Towards meeting the requirements set by experts in the legal field, strong predictions are required to be made from case law. Generally, the computation of high-stakes legal problems requires expensive annotation efforts by domain-knowledge experts. As a result, the cost and complexity of annotating this data makes it difficult to obtain a high number of intermediate labels{\textemdash}for instance, visual or verbal similarity in the case of trademark confusion. However, final judgments are public and thus easy to obtain, and end-to-end text based encoders serve well in making these predictions, which may nonetheless be due to spurious correlations \cite{Bhambhoria2021Investigating}. One of the main objectives of this work is to predict intermediate outcomes. For instance whether trademarks are visually similar{\textemdash}insofar as it constitutes key features for interpretability.
The other main objective is to utilize these intermediate outcomes towards drawing final conclusions for the right reasons i.e. not due to spurious correlations.

 Limited historical case law data reflects the reality across many branches of law, specifically due to the fact that different jurisdictions may have variations in regulations \cite{10.2307/1121857}.  \citet{mosbach2021on} provide a study on fine-tuning models for better generalization on small datasets. In order to establish stronger baselines, we explore and leverage various methods commonly used with deep learning models in order to account for the small amount of annotated data available. Multi-task learning \cite{kendall2018multitask}, albeit beneficial for several tasks which may improve from parameter sharing of neural networks does not benefit our intermediate-label prediction task or final outcome due to the gap between the relations for the two tasks \cite{bingel2017identifying,DBLP:journals/corr/abs-1811-00196}. 
Modern deep neural networks outperform conventional methods when trained on a large amount of data. However, in many high-stakes decision making domains, such as law, leveraging a large amount of data has shown limited performance gains \cite{chalkidis2020legalbert,Bhambhoria2021Investigating}. In fact, models tend to deliver reliable and strong results when trained on high-cost, structured, small amounts of sparsely annotated data \cite{Ustun_2015,rudin2019stop}.
In order to maximize the capabilities of deep learning models, we must use methods such as transfer learning \cite{10.5555/2969033.2969197}, and perform some careful fine-tuning due to a sensitive hyperparameter optimization process. Weak or distant supervision of models is another method by which we can introduce supervision from a large number of noisily labelled data \cite{articletwitter} obtained from heuristics and rules. In this work, we create an augmented dataset which is utilized to provide weak supervision for our models. Unfortunately, naively introducing these noisy samples leads to poor performance. For this reason, we explore adopting strategies such as curriculum learning \cite{10.1145/1553374.1553380,DBLP:journals/corr/abs-2010-13166} to achieve performance gains, while maintaining fidelity to the original model.

The contributions of this work are three-fold:
\begin{itemize}
    \item We are the first to tackle the challenge of risk of confusion for trademark by predicting interpretable intermediate scores which can be utilized to improve the final outcome, i.e. whether there is an infringement of IP law.
    \item We empirically show the effects of utilizing curriculum learning with weakly supervised learning to ensure generalizability, robustness, and stability of our model which can be used for other similar legal decision-making tasks. 
    \item We release the first trademark case law dataset for tackling the challenge of assessing risk of confusion.
\end{itemize}

\section{Related Work}

\begin{figure*}[ht]
\centering
\includegraphics[width=1.6\columnwidth, height=4.5cm]{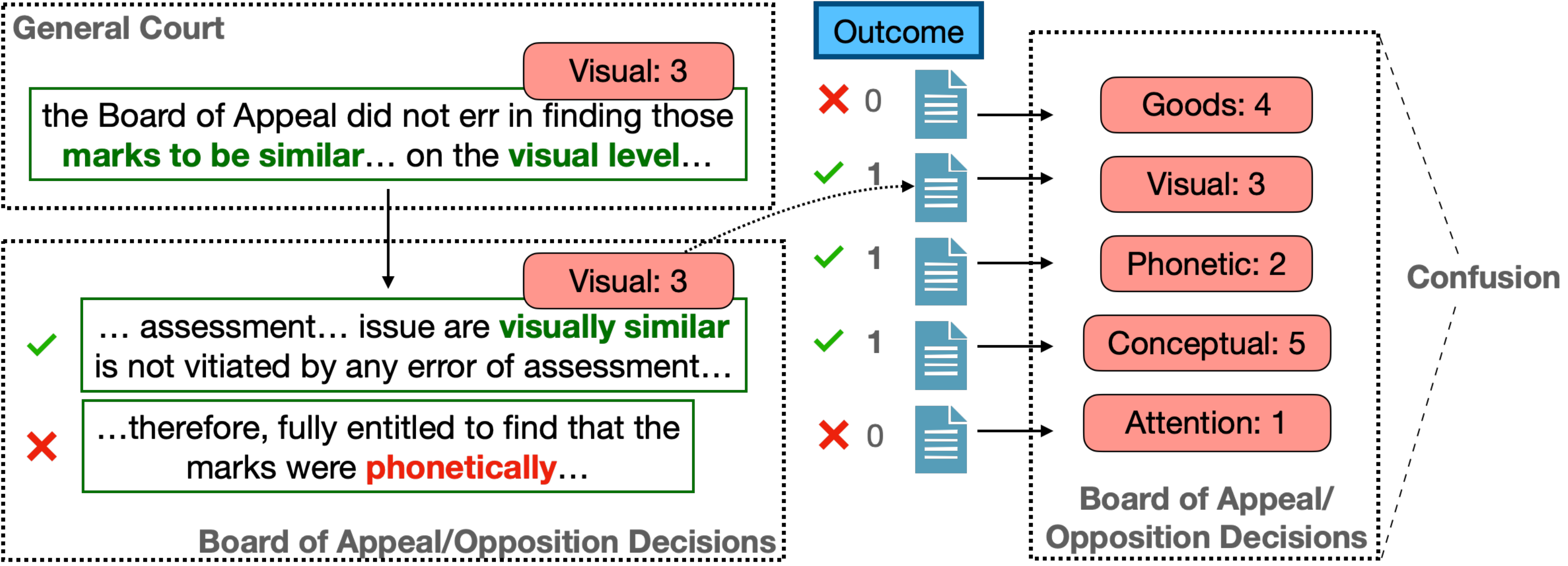} 
\caption{Augmented dataset labels (Board of Appeal/Opposition Decisions) are assigned the intermediate labels from the clean dataset (General Court) which was used for extracting the relevant sentences based on cosine similarity. The outcome of the final judgment (shown as $0$ for ``No Confusion" and $1$ for ``Confusion") is assigned by taking the majority of the publicly available labels from the board of appeal/opposition decision cases.}
\label{fig:data}
\end{figure*}


Recent research has explored blanket terms of explainability and interpretability in great detail. Some works use these terms interchangeably \cite{DBLP:journals/corr/abs-1902-10186,DBLP:journals/corr/abs-1908-04626}, whilst others highlight their intricate differences \cite{DBLP:journals/corr/abs-1806-00069}. In this work, we opt to use the definitions of ``interpretability" as per \citet{DBLP:journals/corr/abs-1806-00069} to highlight the fact that our model as a function is interpretable, and the explanations produced have high fidelity aswell as transparency. Previous work dealt with interpretability by means of introducing post-hoc methods which are model agnostic and may be reliant on perturbations \cite{lime, Ribeiro_Singh_Guestrin_2018} or grounded deeper in axioms \cite{10.5555/3305890.3306024,srinivas2021rethinking}. Other methods explored interpretability on various degrees, i.e. on the level of individual neurons \cite{dhamdhere2018important,bau2020units}, or based on the role of layers \cite{binder2016layerwise}. Another method of providing interpretability is by means of counterfactuals \cite{10.1145/3351095.3372850}. As these methods involve providing explanations for black boxes, they have been met with criticism \cite{rudin2019stop}, in favor of inherently interpretable models \cite{rudin2021interpretable}. 

Several papers focus on interpretability, with a narrower focus on the task of legal judgment prediction trained on large annotated datasets \cite{ye-etal-2018-interpretable,jiang-etal-2018-interpretable,Zhong2020IterativelyQA}. Unlike the present study, past work did not rely on costly and time-consuming annotations by subject-matter experts. For these reasons, these studies do not adequately reflect the real-world scenario for the application of deep learning towards high-stakes decision making. \citet{phang2019sentence} utilized intermediate predictions in order to predict outcomes, although the main motive is not towards providing interpretability, and again, these intermediate predictions do not reflect real-world challenges. Moreover, intermediate labels in the aforementioned work are abundantly available in contrast to the main task.

\citet{wei-zou-2019-eda} explored simple methods to augment data for text classification tasks. However, in high-stakes decision making we must take into consideration data augmentation methods which enables generalization of our model, especially in the legal domain wherein data from different levels of courts may be available. Naively introducing these noisy augmented samples will lead to a reduction in performance. \cite{DBLP:journals/corr/abs-2010-13166} provides a detailed overview of various curriculum learning strategies through which augmented samples can be introduced. In this work we opt for a modified self-paced learning strategy. \citet{wu2021when,zhou2021robust} detailed the scenarios in which noisy samples of data provide benefits towards the training process. In this work, we utilize the traditional method of curriculum learning which introduces noisy samples from easy-to-hard for training.

\section{Dataset}


\begin{table}[t]
\centering
\begin{tabular}{|l|l|l|}
\hline
\textbf{Subset}    & \textbf{Num. Cases} & \textbf{Avg. Char/$S$} \\ \hline
\textbf{Clean}     & 525                 & 1081                              \\ \hline
\textbf{Augmented} & 2852                & 941                               \\ \hline
\end{tabular}
\caption{Clean subset represents cases annotated by lawyers, obtained from the CJEU. Augmented subset represents cases which are extracted from the Board of Appeal/Opposition Decisions. $S$ denotes the input text comprising of factors which influence the likelihood of confusion.}
\label{subset}
\end{table}


The datasets we created for the purpose of this study are comprised of two parts{\textemdash}firstly, cases obtained from the General Court, a constituent of the Court of Justice of the European Union (CJEU)\footnote{https://curia.europa.eu/}. The CJEU is the highest judicial authority in the field of trademark in the European Union. The size of the data sample is significant from a legal standpoint insofar as we were able to extract a large pool of decisions related to trademark confusion. As a result, the CJEU body of case law constitutes a meaningful and relevant source of training data for our models. The cases obtained from the CJEU were annotated by a team of lawyers with a training in trademark law. An annotation guideline was designed by legal professionals, taking into consideration the various factors which are involved in deciding the outcome of a case. Despite the guideline comprising of 32 fields, a condensed set of 5 factors were deemed important towards analyzing the risk of confusion for the likelihood of trademarks, shown in Table \ref{annotate:guidelines} of the Appendix. These factors are ``Similarity of Goods \& Services”, ``Visual Similarity”, ``Conceptual Similarity”, ``Phonetic Similarity”, and ``Level of Attention”. Few cases are more nuanced and require ``split decisions” to be made for annotations. We exclude the consideration of these outlier cases in this work. Due to the presence of missing values in our annotated cases under the selected important features, we have a set of 426 cases, from which 213 samples are reserved for training, 53 for validation, and 160 for testing. All cases are extracted from 2008 onwards, as major changes\footnote{For the sake of consistency, we have annotated cases decided after Regulation (EC) No. 207/2009, which brought forward various changes in trademark law.} in trademark law have been made since. The resulting cases are hence comprised of 525 annotated documents in its entirety. Secondly, we weakly labelled an additional 2852 cases from a larger pool of ``Opposition decisions”, and ``Board of Appeal decisions” \footnote{https://euipo.europa.eu/eSearchCLW/}. The average length of these document are comparable, as shown in Table \ref{subset}. These additional cases are similar in terms of semantics to our expert-annotated cases, with the exception that from a legal perspective, they were decided by a lower authority and thus do not bear the same legal value as the cases of the CJEU. We refer to these cases as ``augmented data”, the entirety of which is utilized for training under different settings. The detailled construction process of these datasets will be elaborated below.

\subsection{Obtaining Expert-Annotations}

Factors that are relevant to determine the existence of confusion among trademarks were drawn from the Regulation (EC) No 207/2009), the foundational legislation in the field of trademark. Lawyers were trained on identifying these factors. Sentences were extracted corresponding to these important features, and were labelled on a predefined scale. In this work, the resulting dataset is regarded as a ``clean dataset”. Next, a control-group study was created, wherein the independent variable was set to be the final outcome i.e. confusion or no confusion. Factors influencing the judgment were presented to the control group consisting of lawyers. The participants were asked to determine the outcome when presented with only the labelled factors. A human-annotated Macro F1 score of $\approx 86\%$ was achieved in determining confusion for this task. The factors responsible for the inability in obtaining perfect assessment are as follows{\textemdash}1) minor concluding statements by courts may shift the judgment, 2) courts may have paid more attention to specific features in a few cases, 3) human annotator error, 4) case contained varying claims, 5) split-decisions in cases. Participants involved in this control group were also asked to identify rules involved in assessing the likelihood of confusion. These rules were utilized to establish the labels for the cases from ``Opposition Decisions”, and ``Board of Appeal Decisions”, and are detailed in Table \ref{rules:guidelines} of the Appendix. 

\subsection{Obtaining Augmented Cases}

The capability of a simple process to obtain augmentation for this task is possible due to the repetitiveness in trademark case law. We first retrieve $\approx 52000$ cases from the ``Opposition Decisions”, and ``Board of Appeal Decisions” dating from 1998 to 2020. This serves as a large pool of initial cases, from which we follow the extraction, randomization, filtering, and assigning steps, shown in Figure \ref{fig:data} detailed as follows:

\begin{enumerate}

\item \textbf{Extraction:} We utilize cases from our annotated clean training dataset and create TF-IDF matrices for all sentences. We also repeat this process for the lower court (augmented, pre-filtered) cases, of which we have a larger pool of cases. Then, we calculate cosine similarity and extract the top sentences corresponding to the sentences obtained from our training set.

\item \textbf{Randomization}: As extracted sentences may correspond to one of five features (Similarity of Goods \& Services, Visual Similarity, Conceptual Similarity, Phonetic Similarity, or Level of Attention), we randomly concatenate sentences corresponding to these features to create concluding paragraphs, which albeit noisy, contain a large number of samples.  

\item \textbf{Filtering:} By utilizing a few basic keywords which are known to be present in sentences attributing to confusion (e.g. `conceptually` for `Conceptual Similarity`), we obtain basic filtered textual inputs. We then utilize the rules devised by the control group study to further filter our augmented samples.

\item \textbf{Assigning:}
We utilize the annotated (clean) samples to assign pseudo-labels to these augmented samples by maintaining the original labels from which they were extracted for intermediate labels. We also make use of the final outcome to assign a label for the final judgment of the case by consideration of the majority number of samples attributing to a label of ``Confusion" or ``No Confusion". For example, if a majority of the cases from which the statements are derived from consist of the final judgment of confusion, the pseudo-label for the document will also be confusion. In this work, the resulting cases obtained from this stage will be referred to as our ``augmented dataset”.

\end{enumerate}

\section{Methodology}

\begin{figure}[t]
\centering
\includegraphics[scale=0.35]{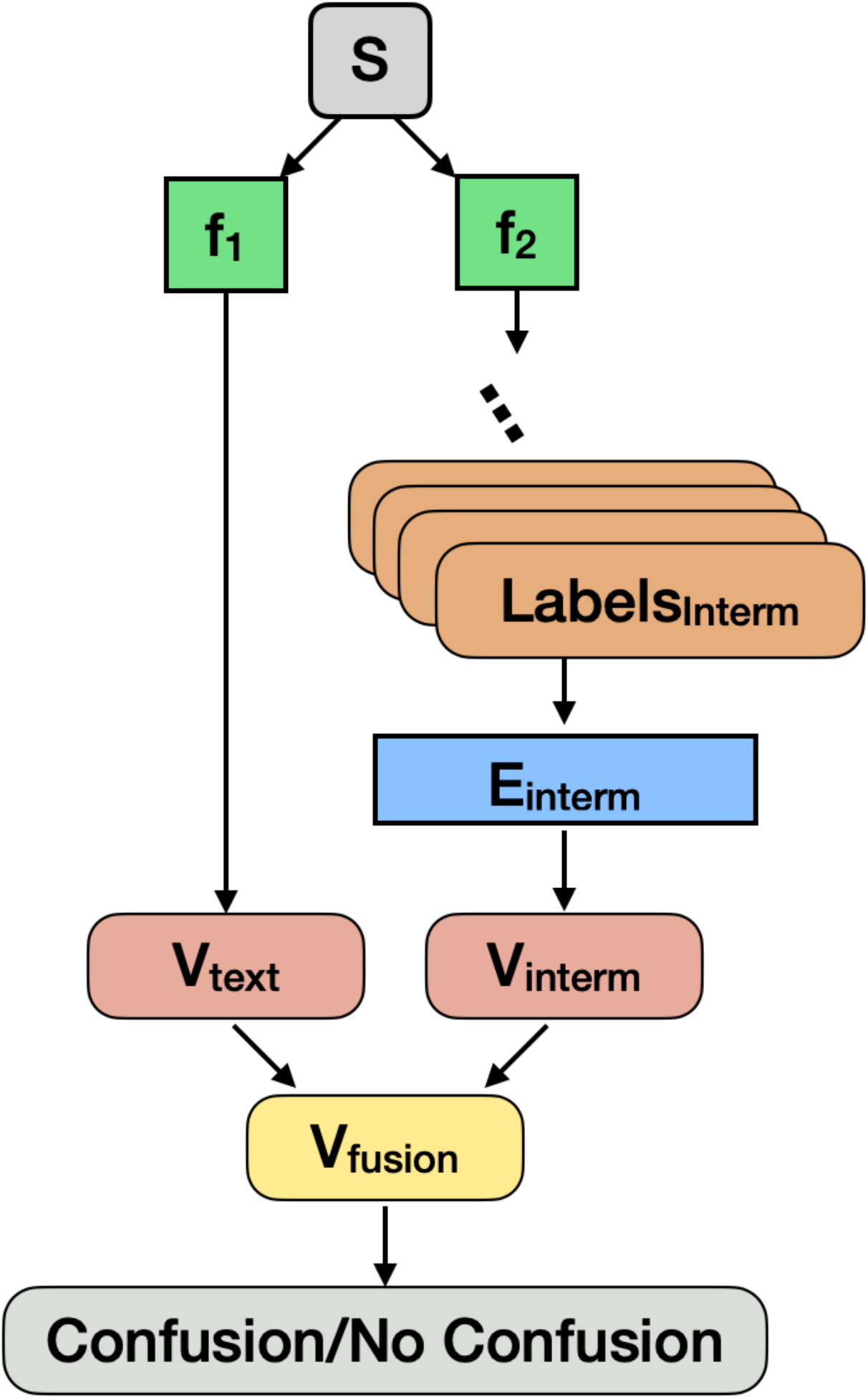} 
\caption{Our proposed model which takes into consideration intermediate labels for interpretability, and addresses the challenge of decision-making due to spurious correlations. $S$ denotes the input excerpts extracted from case law. $f_1$ and $f_2$ are initialized by fine-tuning on independent loss functions. $V_{interm}$ is produced by first calculating intermediate labels, after which an encoder denoted by $E_{interm}$ enables us to fuse it with the representation obtained from $f_{1}$. $V_{text}$ and $V_{interm}$ are fused to create $V_{fusion}$.}
\label{fusionmodel}
\end{figure}

The motivation for designing the model in this work is to introduce interpretability in a model-agnostic setting which can be utilized for legal decision-making tasks.
As shown in Figure \ref{fusionmodel}, this is done by introducing an abstracted layer derived from text, which we call the components of as ``intermediate labels". By predicting these labels, we are able to induce interpretability in our model. Furthermore, by utilizing these intermediate labels to make judgments of the final outcome, we enable the model to make decisions for the right reasons. In other words, we tackle the challenge of spurious correlations wherein decisions may be made correctly due to incorrect reasons by introducing these intermediate labels for producing the outcome. As our ``clean dataset" contains only a few samples, we utilize our ``augmented dataset" to promote generalization of our model through a curriculum learning strategy which is detailed in Figure \ref{curriculumfigure}.

\subsection{Baselines}

\subsubsection{End-to-End} 
Transformer-based models such as BERT \cite{devlin-etal-2019-bert} have shown strong performance on various text classification tasks. We utilize RoBERTa \cite{DBLP:journals/corr/abs-1907-11692} for the task of predicting the likelihood of confusion for trademarks. This model is tested on the various splits of training data{\textemdash}1) clean data, 2) augmented data, and 3) mix data which is simply a combination of the shuffled clean, and augmented data. The experiments under these settings are performed separately for both, likelihood of confusion classification task, and for intermediate-label prediction regression task as detailed in Figure \ref{fig-etebaseline} of the Appendix. As a consequence of this disjoint training, the two tasks are unable to benefit each other, and hence decisions for final labels may be made due to spurious correlations.

\subsubsection{Multi-Task}

As our task is targeted to tackle the optimization of intermediate labels, and the prediction of likelihood of confusion, it is reasonable to consider a multi-task structure wherein parameters are shared across two identical encoders as shown in Figure \ref{fig-mtlbaseline} of the Appendix. As with our end-to-end model, we use RoBERTa \cite{DBLP:journals/corr/abs-1907-11692} for the multi-task setting in order to ensure consistency across our results. However, we still face the challenge of spurious correlations due to the inability of this method to build a connection between the two tasks.

\subsection{Fusion Model}
To overcome the aforementioned weaknesses in the baseline methods, wherein decisions may be made due to spurious correlations, and to address the need for interpretability, we propose a fusion model shown in Figure \ref{fusionmodel}.

For a given set, $S = \{s_1, s_2,...,s_n\}$ $\forall {s_i}\, \in S^{\prime}$ we have $S^{\prime}$ which corresponds to factors in the form of text influencing the likelihood of confusion either extracted manually in the case of the clean dataset, or automatically obained in the case of the augmented dataset. 

We utilize an encoder, $f_1$ which is initialized from the best performing checkpoint of the end-to-end model for the binary prediction task to determine the likelihood of confusion in a case. This is done by utilizing cross entropy loss shown in Equation \ref{eqn-1} wherein $m$ are the number of samples, $y_{i}$ is the target label, and $\hat{y}_{i}$ is the output obtained from $f_1(S)$. 

\begin{equation}
    L=-\frac{1}{m} \sum_{i=1}^{m} y_{i} \cdot \log \left(\hat{y}_{i}\right)
    \label{eqn-1}
\end{equation}

For the intermediate task, encoder $f_2$ is initialized by a similar method after being fine-tuned by optimizing Equation \ref{eqn-2}, for the value of variables detailed in Equation \ref{eqn-2.5}. We use Smooth L1 Loss here as it accounts for outliers, and leads to better performance in lieu of utilizing mean absolute error, or mean standard error loss. The unreduced loss is given by:

\begin{equation}\label{eqn-2}
    L=\left\{l_{1}, \ldots, l_{I}\right\}^{T},
\end{equation}
where $l_{i}$ can be formatted as:
\begin{equation}\label{eqn-2.5}
    l_{i}=\left\{\begin{array}{ll}0.5\left(\hat{y}_{i}-y_{i}\right)^{2} / \beta , & \text { if }\left|\hat{y}_{i}-y_{i}\right|<\beta \\ \left|\hat{y}_{i}-y_{i}\right|-0.5 * \beta , & \text { otherwise }\end{array}\right.
\end{equation}
where $\hat{y_{i}}$ and $y_{i}$ denote the output of $f_2(S)$ and the target respectively. $\beta$ is a hyperparameter, wherein if $\beta \to 0$, smooth L1 Loss will converge to L1 Loss, and if $\beta \to +\infty$, will converge to a constant 0 loss. We set $\beta$ to 1, to account for outliers.

A feedforward layer which applies a linear transformation is utilized to obtain the label values, after which we utilize an embedding layer. A transformer block made up of self-attention as shown in Equation \ref{eqn-3} and a feedforward network is then utilized as an encoder. A final feedforward connection is used to obtain the representation for the intermediate labels. We do not directly consider the hidden representation of $f_{2}$ here due to the fact that we require our model to produce the intermediate labels, a property which is not obtainable otherwise.

\begin{equation}
    Attention = \operatorname{softmax}\left(\frac{Q K^{T}}{\sqrt{d_{k}}}\right) V
    \label{eqn-3}
\end{equation}
where $Q$ is a query matrix, i.e. embedding of $s_{i}$. $K$ are the keys, i.e. the embedding of all words in $S$. $V$ are values, and $\sqrt{d_{k}}$ is the dimension of $K$, by which scaling is performed as per \citet{DBLP:journals/corr/VaswaniSPUJGKP17}.

The fusion layer is a modified version of \citet{cai-etal-2019-multi} and is detailed in Equation \ref{eqn-4}.

\begin{equation}
    V_{Fusion}=s1*V_{text}+s2*V_{interm}
    \label{eqn-4}
\end{equation}
where $s1$ and $s2$ are scores calculated by using hyperbolic tangent function denoted in Equation \ref{eqn-5} after a feedforward layer stemming from the concatenated vector representation of $V_{text}$ with $V_{interm}$, and conversely $V_{interm}$ with $V_{text}$ respectively: 
\begin{equation}
    \tanh (x)=\frac{e^{x}-e^{-x}}{e^{x}+e^{-x}}
    \label{eqn-5}
\end{equation}
where $x$ can be set to the concatenated vector representations to obtain $s1$, and $s2$ respectively.

\subsection{Curriculum learning}

\begin{figure}[ht]
    \centering
    \includegraphics[scale=0.32]{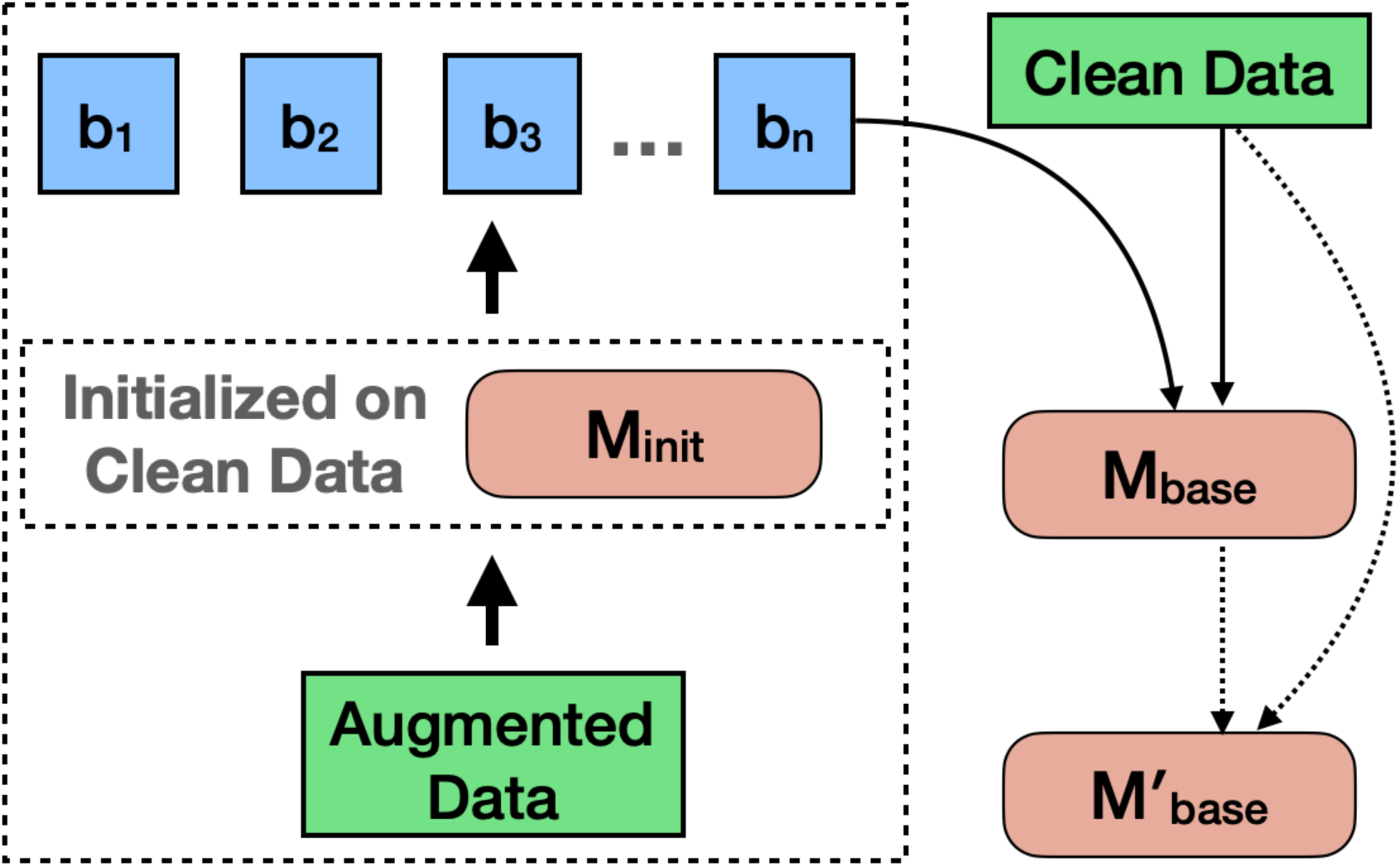}
    \caption{Curriculum learning strategy by which augmented data is utilized to improve predictive capabilities of a model.}
    \label{curriculumfigure}
\end{figure}

As our augmented dataset contains several noisy samples, it is natural to consider a structured method of introducing samples in order of easy-to-difficult for training our model. We adopt a curriculum learning strategy to be utilized in introduction of samples wherein the base classifier is utilized to set a threshold for difficulty on the basis of softmax probabilities as detailed in Figure \ref{curriculumfigure}. The model, $M_{init}$ is trained on only clean data. $M_{base}$ is an untrained encoder with the same architecture as $M_{init}$. By dividing the augmented training dataset into bins, $B = \{b_{1}, b_{2},..., b_{n}\}$, we select the bins containing confident samples to be introduced as training samples with clean data for $M_{base}$. The updated $M_{base}^{\prime}$ is further trained on only clean data and serves as our model for the curriculum learning strategy.

\section{Experimental Setup}

In all experiments, the $f_{1}$ and $f_{2}$ architectures utilized are RoBERTa-base models. In our model, the embedding layer in $E_{interm}$ has a dimension of 300, and is arbitrarily set. The transformer block in $E_{interm}$ for the intermediate labels consists of 6 layers with 6 heads, and a feedforward dimension of 512.
In our experiments, we utilize a batch size of 8 for all experiments on a single RTX 2080Ti machine with 11 GB of memory, taking into consideration the limited computational resources practioners in the legal field may have. For curriculum learning, we arbitrarily set the number of bins to be 3 wherein the first bin contains the most confident samples with a probability threshold of $\geq 0.99$, the second bin contains the remaining correct samples, and the final bin contains all samples from the mixed dataset for training which were not contained within the previous bins. For the calibration experiments, we arbitrarily set the number of bins to be 5. 
PyTorch-based libraries are utilized for all experiments.

\section{Results and Discussion}

Our objective with the experiments is to demonstrate with strong empirical evidence that our model is capable of producing stable, reliable results with excerpts obtained from case law. The results are shown on our carefully curated dataset for the likelihood of confusion for trademarks in Table \ref{results-1}. However, the findings in this work can easily be utilized for other similar legal decision-making tasks. 

End-to-end models may suffer from issues such as making decisions due to spurious correlation, as they are in no way constrained by the intermediate labels. On the other hand, the multi-task baseline, and our model is capable of utilizing the intermediate labels to make decisions. As shown in Table \ref{results-1}, there is no benefit in utilizing the end-to-end model as a predictive solution for this task. Firstly, the performance is much worse than our model, and the multi-task model for the clean dataset. For the mixed dataset, it is unable to benefit from training on the larger number of noisy samples. Secondly, the standard deviation between different runs is very high for random starts of the model for training. The use-case of the end-to-end model is only as initializing $M_{base}$ for the curriculum learning strategy. 

The multi-task model, despite incorporating the intermediate labels during training, is unable to take advantage of these labels in improving the performance of likelihood of confusion. For the clean dataset, the performance is comparable to that of the end-to-end model. On the other hand, for the mixed dataset, the performance is slightly worse than the end-to-end model. Results from both data splits show high standard deviation, and hence unstable results between runs. However, as this model utilizes a shared encoder wherein the weights are updated from both, intermediate, and final outcome prediction tasks, it is a step in the right direction towards making decisions for the correct reasons, and not due to spurious correlations.

\begin{table}[t]
\centering
\begin{tabular}{|l|l|l|}
\hline
\multicolumn{1}{|c|}{\multirow{2}{*}{\textbf{Model}}} & \multicolumn{2}{c|}{\textbf{Data Split}}                                \\ \cline{2-3} 
\multicolumn{1}{|c|}{}                                & \multicolumn{1}{c|}{\textbf{Clean}} & \multicolumn{1}{c|}{\textbf{Mix}} \\ \hline
Multi-Task                                   & $75.52 \pm 6.13$                               & $60.51 \pm 7.34$                             \\ \hline
End-to-End                                   & $75.22 \pm 6.89$                               & $62.47 \pm 7.47$                             \\ \hline
Ours                                         & $\textbf{79.16} \pm 2.34$                               & $65.60 \pm 3.01$                             \\ \hline
Ours   + curric.                             & -                                   & $\textbf{83.05 }\pm 2.54$                             \\ \hline
\end{tabular}
\caption{Macro F1 scores reported for the binary prediction task of likelihood of confusion ($\pm$ standard deviation). Multi-task denotes fine-tuned jointly-trained roberta-base encoders. End-to-End denotes a fine-tuned roberta-base model. Results shown are averaged across 5 runs. The data split ``Mix" is the randomly shuffled concatenation of the clean and augmented datasets. ``Curric." is an abbreviation for our curriculum learning straegy shown in Figure \ref{curriculumfigure}.}
\label{results-1}
\end{table}

\begin{table}[t]
\centering
\begin{tabular}{|c|c|c|}
\hline
\textbf{Model} & \textbf{MAE} & \textbf{MSE} \\ \hline
Random         &      $1.81 \pm 0.03$        &      $5.09 \pm 0.12$        \\ \hline
\multicolumn{3}{|c|}{\textbf{Clean}}         \\ \hline
Multi-Task     &       $0.75 \pm 0.03$       &      $1.05 \pm 0.06$        \\ \hline
Ours     &       $\textbf{0.75} \pm 0.03$       &      $\textbf{0.99} \pm 0.07$         \\ \hline
\multicolumn{3}{|c|}{\textbf{Mix}}           \\ \hline
Multi-Task     &        $0.84 \pm 0.04$      &      $1.27 \pm 0.13$         \\ \hline
Ours     &        $0.83 \pm 0.05$      &      $1.24 \pm 0.13$        \\ \hline
\end{tabular}
\caption{Mean Absolute Error (MAE) and Mean Standard Error(MSE) reported for regression task of predicting intermediate labels (lower value is better). The loss function utilized for training is Smooth L1 Loss which is more stable for training on lesser number of samples.}
\label{results-2}
\end{table}

Our model is able to incorporate the intermediate labels, much like the multi-task model, and hence also tackles the challenge of decisions being made due to spurious correlations. However, the major benefit is seen on both splits of data, wherein the standard deviation is drastically reduced between runs, hence delivering stable results. Furthermore, the performance shows great improvement under the clean dataset data split when compared with the end-to-end, and multi-task models. Nonetheless, the benefits of utilizing the augmented dataset are only apparent when we utilize our curriculum learning strategy to introduce these samples from the order of easy-to-difficult. Here, we observe a drastic improvement over the previous ``Mix" data splits, and better performance than our previously best-performing clean data-split from our model.

\begin{figure}
    \centering
    \includegraphics[width=0.95\columnwidth]{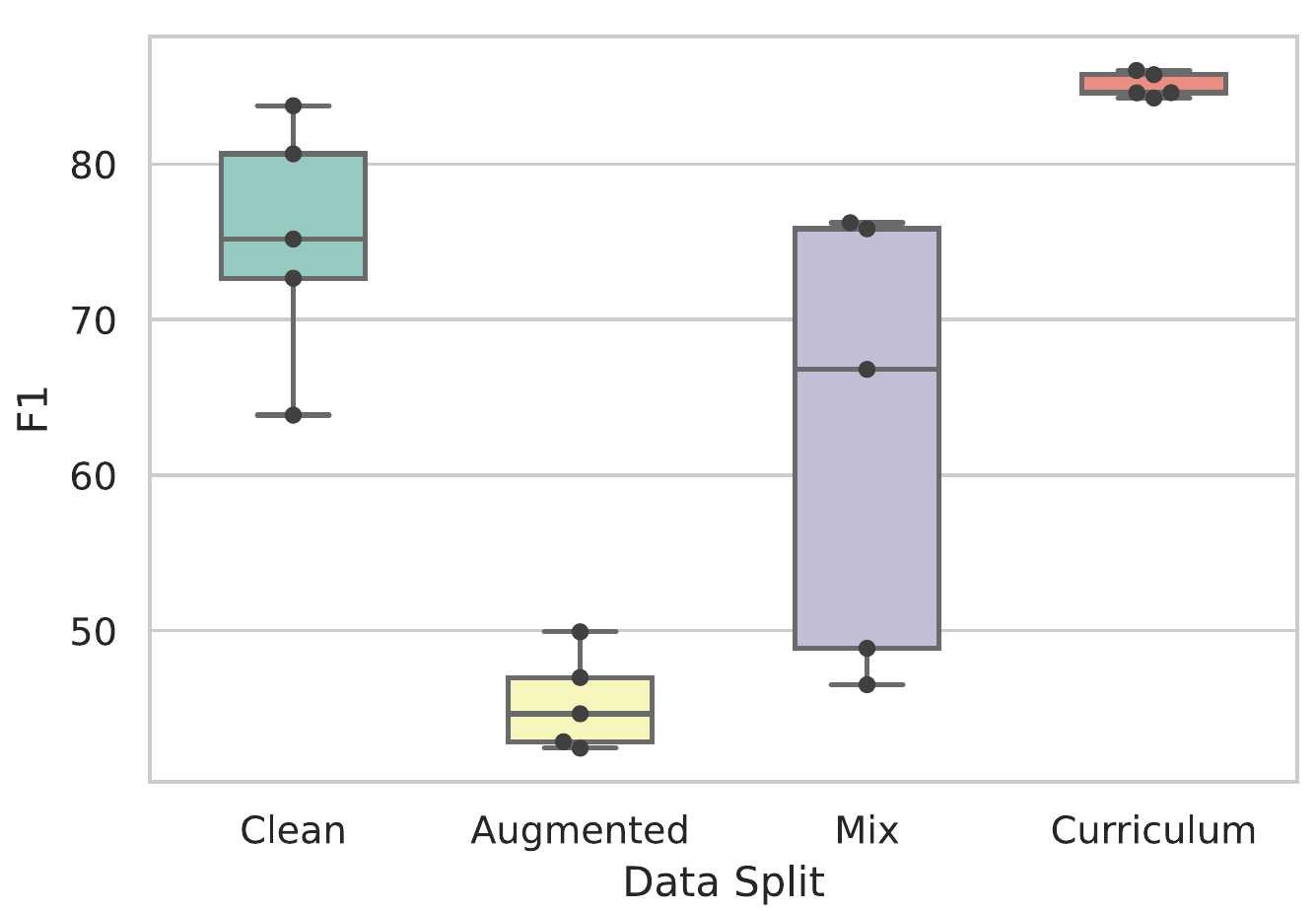}
    \caption{Stability and improved performance of our model is showcased when utilizing curriculum learning. Naively utilizing the mixed dataset leads to poor performance.}
    \label{box-plot}
\end{figure}

As computational resources available to legal decision-makers may be limited, it is crucial that stability of models is obtained from deep neural network architectures without excessive efforts for fine-tuning. From Figure \ref{box-plot}, we observe that the results when utilizing the clean dataset are unstable, and requires multiple runs to achieve desirable performance. For the augmented dataset alone, due to dataset shift between the augmented dataset and clean dataset, very poor performance is showcased. This poor performance is expected, and accentuates the requirement of a few clean samples during the training process. The benefit of utilization of the mixed dataset is only apparent when we introduce it by means of curriculum learning, from which we are able to achieve stable and strong performance.


For the intermediate task, we utilize Smooth L1 Loss, as it leads to stable training with small batches which is a product of our small datasets for training. We report both MAE and MSE as metrics to showcase the best performing model for intermediate-label predictions in Table \ref{results-2}. Expectedly, randomly setting these intermediate labels shows poor performance. As with our models for prediction of likelihood of confusion as the ground truth, we observe no benefit in utilizing the multi-task model for predicting the intermediate labels. Furthermore, we omit the results of curriculum learning strategies in predicting the intermediate labels as they did not provide any benefit for intermediate-label prediction performance due to the synthetic nature of the augmented dataset. Utilizing the mixed dataset also did not show any improvement in performance. 

Through experimental results, we observe no benefit in initializing our fusion-based model from models which are not finetuned. Instead, we utilize the best fine-tuned checkpoints for each respective data split. Inclusion of these intermediate labels provide transparency of the fusion model, wherein results are explainable to a degree that they can be utilized by lawyers to assess the risk of confusion for trademarks.






\begin{table}[ht]
\centering
\begin{tabular}{|p{0.42\columnwidth}|p{0.22\columnwidth}|}
\hline
\textbf{Model} & \textbf{ECE} \\ \hline
End-to-End     &          0.1452         \\ \hline
Multi-task     &           0.1520        \\ \hline
Ours           &           \textbf{0.0733}        \\ \hline
\end{tabular}
\caption{Expected Calibration Error (ECE) of our fusion model out-performs the baselines (lower values are better).}
\label{calibration-results}
\end{table}

\subsection{Calibration}

As first shown in \citet{10.5555/2888116.2888120}, and later compared in \citet{DBLP:journals/corr/GuoPSW17} for neural networks, we utilize Expected Calibration Error (ECE) to showcase a statistic regarding the calibration comparison of our models on making the final conclusions for assessment of likelihood of confusion. The importance of calibration metrics of models in high-stakes decision making is to ensure that the confidence is taken into consideration for the choice of a model, and this selection is not dependant only on the predictions.

By creating $M$ bins, where $B_{m}$ are samples belonging to bins in confidence intervals $I_{m}=\left(\frac{m-1}{M}, \frac{m}{M}\right]$, we can calculate the ECE value from Equation \ref{calib-3} which are dependent on the predictions as in Equation \ref{calib-1} and confidence of these values shown in Equation \ref{calib-2}.

\begin{equation}
    \mathrm{ECE}=\sum_{m=1}^{M} \frac{\left|B_{m}\right|}{n}\left|\operatorname{acc}\left(B_{m}\right)-\operatorname{conf}\left(B_{m}\right)\right|
    \label{calib-3}
\end{equation}

\begin{equation}
    \operatorname{acc}\left(B_{m}\right)=\frac{1}{\left|B_{m}\right|} \sum_{i \in B_{m}} \mathbf{1}\left(\hat{y}_{i}=y_{i}\right)
    \label{calib-1}
\end{equation}
where $\hat{y}_{i}$ denotes the model output, and $y_{i}$ denotes the ground truth.

\begin{equation}
    \operatorname{conf}\left(B_{m}\right)=\frac{1}{\left|B_{m}\right|} \sum_{i \in B_{m}} \hat{p}_{i}
    \label{calib-2}
\end{equation}
where $\hat{p}_{i}$ denotes the probabilities of the predictions.

The results shown in Table \ref{calibration-results} suggest that our model is significantly well-calibrated in constrast to end-to-end, and multi-task models.

\section{Conclusion}
In this work, we explore the application of deep learning towards the task of predicting the likelihood of confusion for trademarks which can easily be extended to other branches of law. We address the data limitations by creation of an expertly-tailored augmented dataset. We observe that training on these cases produces stable results only when coupled with a curriculum learning strategy. We also address the challenge of interpretability by introducing intermediate labels{\textemdash}such as predicting whether trademarks are visually similar, which in turn acts as an interpretable mechanism to constrain decisions being made due to spurious correlations. We show that our fusion model outperforms baseline methods by not only producing improvement in the macro-F1 metric, but by also improving on the ability to predict intermediate labels. Finally, we show a drastic improvement of calibration of our model in comparison to the baselines by reporting the Expected Calibration Error.

\section*{Acknowledgements}
We acknowledge the support of the Government of Canada’s New Frontiers in Research Fund (NFRF), [NFRFE-2018-00715].

\bibliography{aaai22}




\appendix
\clearpage
\section{Appendix}
\label{sec:appendix}

\subsection{Annotation Guidelines}

The annotation guidelines which are provided to lawyers by legal professionals are shown in Table \ref{annotate:guidelines}. Lawyers are required to assess cases they are presented with, parse sentences which correspond to the ``Feature" column, and extract these sentences. They are also required to assign values as per the ``Data Input" column by utilizing the ``Guideline". Entries shown in Table \ref{annotate:guidelines} only contain the intermediate labels, and final judgment for assessment of confusion which are utilized for our experiments out of a total of 32 factors. This list is not comprehensive, and other factors which are not shown are due to the reason they are either{\textemdash}(1) less contributing factors towards the likelihood of confusion such as the ``Difference in a Distinctive Element", for which the data input would be binary indicating the presence or absence of similar distinctive elements between two trademarks\footnote{EUIPO guidelines, section 2, chapter 4 ``A difference in a distinctive element tends to decrease the degree of similarity". https://guidelines.euipo.europa.eu/1803468/1789398/trade-mark-guidelines/}, or (2) Irrelevant towards decision making, e.g. ``Chamber", or ``Rapporteur".
In addition to these factors, to maintain consistency between annotations, we also devise general guidelines as below:
\begin{itemize}
    \item When judges do not take into account a variable that might otherwise be relevant, it is entered as ``Not Considered". 
    \item If a variable does not apply to the particular fact pattern of a case, ``Not Applicable" is entered. 
    \item For outlier data entry, text is entered exactly as it appears, and a note is made in a separate column.
    \item Cases are split when there are two or more considerations under one category e.g. multiple goods and services are covered by both or one mark. The analysis splits as a result the two or more possible entries under that category.
\end{itemize}

\subsection{Control Group Study Guidelines}

As a part of a control group study, lawyers were presented with intermediate labels, and required to assess the likelihood of confusion of a case. Results from two independent lawyers are showcased in Table \ref{rules:guidelines} as per the ``Annotator \#" column. Rules were established based on these decisions and utilized to assess the ability to determine the final outcome. These rules were also used to filter cases in the creation process of the ``augmented dataset".

\subsection{Baseline Architectures}

\subsubsection{End-to-End}

\begin{figure}[ht]
\centering
\includegraphics[width=0.6\columnwidth]{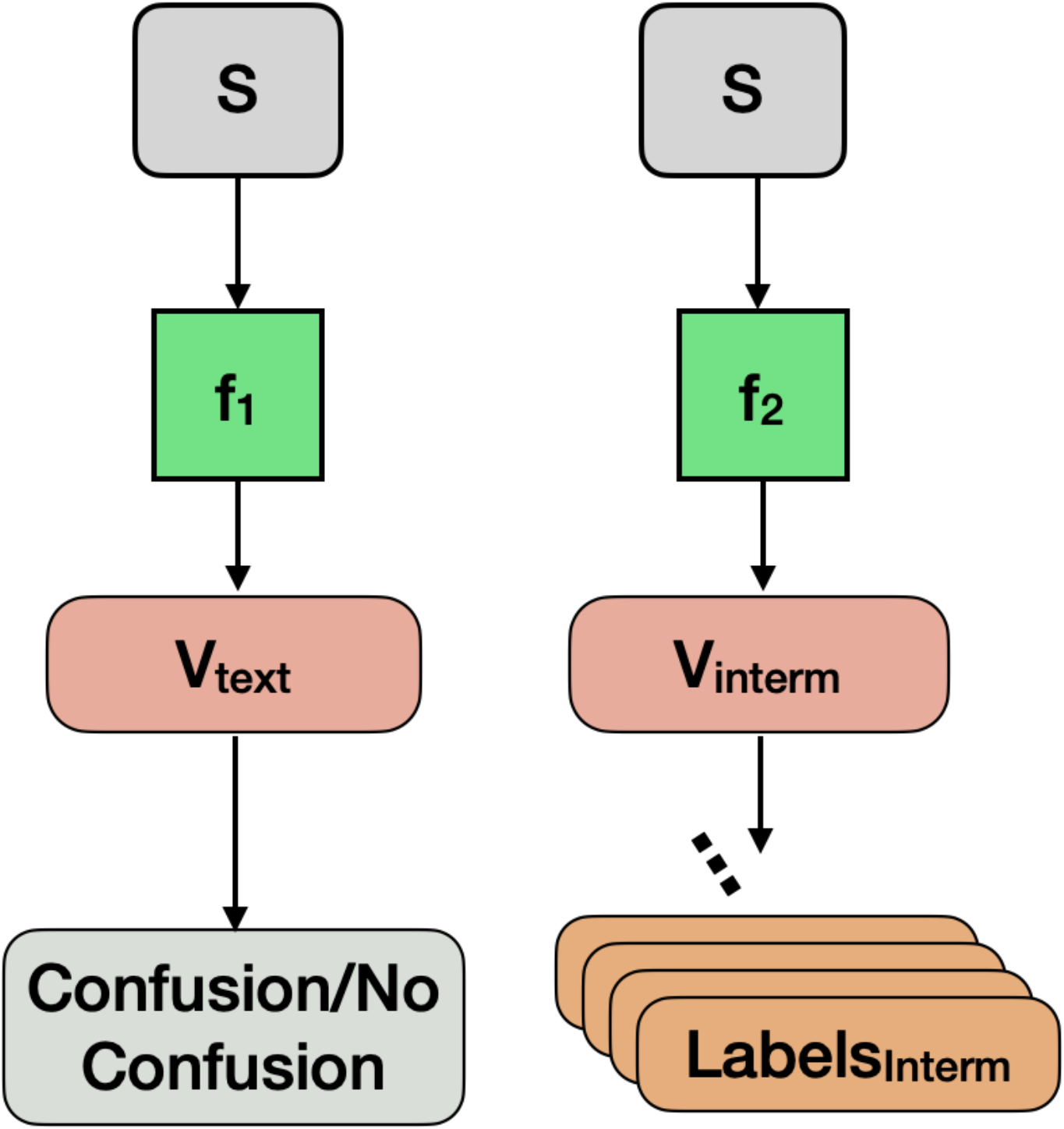} 
\caption{End-to-end model architecture.}
\label{fig-etebaseline}
\end{figure}

End-to-end model shown in Figure \ref{fig-etebaseline} involves disjoint training for prediction of the likelihood of confusion, and intermediate labels. $f1$, and $f2$ here denote encoders with RoBERTa-base architectures, and a feedforward layer with output dimensions of 2, and 5 respectively.

\begin{figure}[ht]
\centering
\includegraphics[width=0.6\columnwidth]{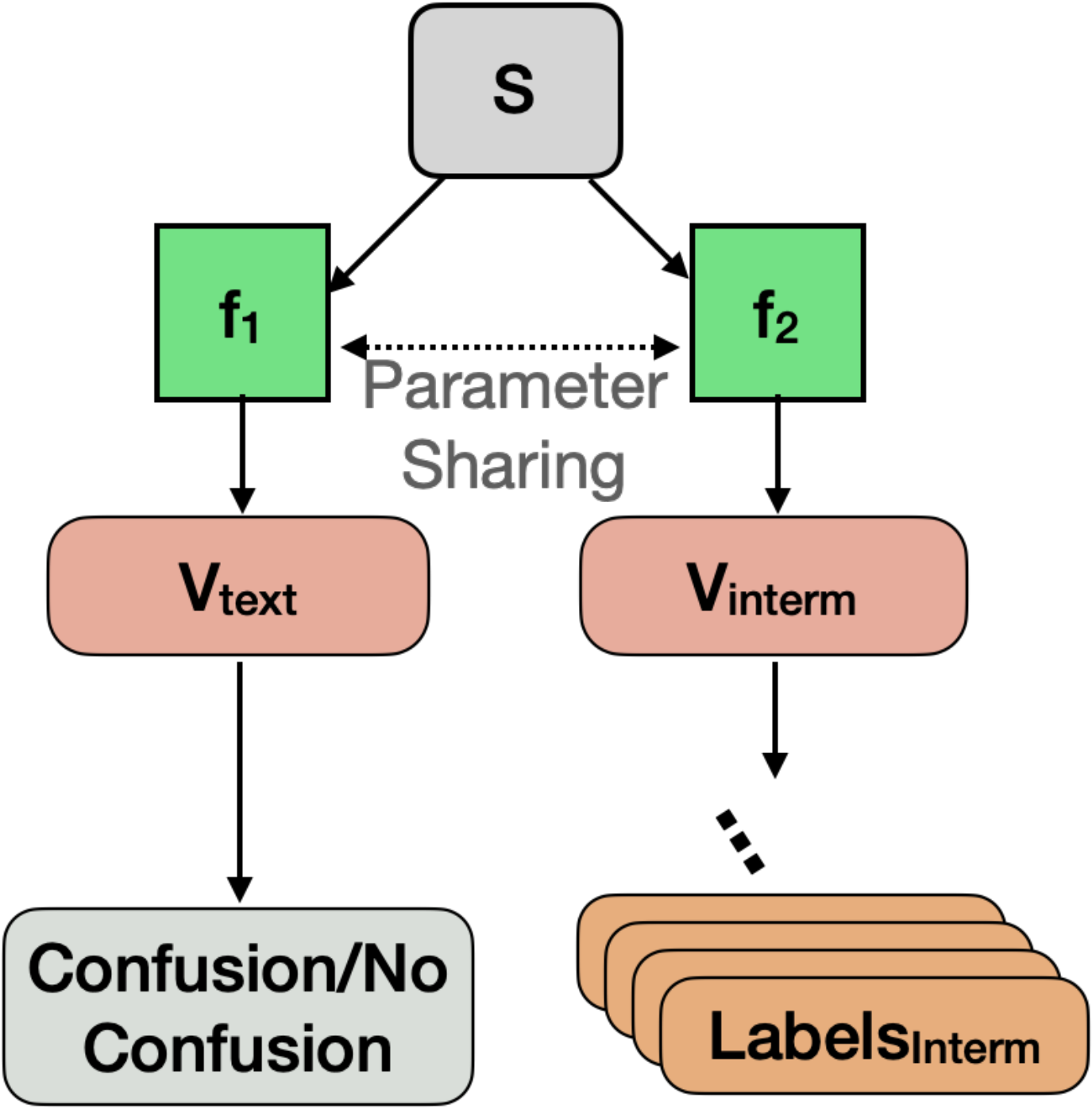} 
\caption{Multi-task learning-based model architecture.}
\label{fig-mtlbaseline}
\end{figure}

\subsubsection{Multi-task Learning}
Multi-task learning-based model shown in Figure \ref{fig-mtlbaseline} involves parameter sharing with $f1$, and $f2$, wherein the loss functions utilized are cross-entropy for the likelihood of confusion task, and Smooth L1 loss for the prediction of intermediate labels. $f1$, and $f2$ have RoBERTa-base architectures.

\begin{table*}[hb]
\centering
\begin{tabular}{|l|l|l|}
\hline
\textbf{Feature}                                                             & \textbf{Data Input}                                                                                                                           & \textbf{Guideline}                                                                                                                                                                                                                                                                                                                                        \\ \hline
\begin{tabular}[c]{@{}l@{}}SIMILARITY OF \\ GOOD(S)/SERVICE(S)\end{tabular} & \begin{tabular}[c]{@{}l@{}}identical (5)\\ high similarity (4)\\ similar (3)\\ low similarity (2)\\ dissimilar (1)\end{tabular}               & \begin{tabular}[c]{@{}l@{}}The scale is based on words used to describe similarity of goods and \\ services in the relevant case law.\end{tabular}                                                                                                                                                                                                        \\ \hline
\begin{tabular}[c]{@{}l@{}}VISUAL \\ SIMILARITY\end{tabular}                & \begin{tabular}[c]{@{}l@{}}identical (5)\\ high similarity (4)\\ similar (3)  \\ low similarity (2)\\ dissimilar (1)\end{tabular}             & \begin{tabular}[c]{@{}l@{}}The scale is based on assessments made in the relevant case law. The scale \\is numerical because in some cases, analysts will need to input the average \\ between two numbers. For example, if the court states that the degree of \\ similarity varies between average and high, then that would be 3.5.\end{tabular} \\ \hline
\begin{tabular}[c]{@{}l@{}}PHONETIC \\ SIMILARITY\end{tabular}              & \begin{tabular}[c]{@{}l@{}}identical (5)\\ high similarity (4)\\ similar (3)\\ low similarity (2)\\ dissimilar (1)\\ neutral (0)\end{tabular} & \begin{tabular}[c]{@{}l@{}}Input neutral (0) if there are no concepts that can be  associated with the\\ marks.  Judges will generally state if that is the situation. \\ Otherwise, same as above.\end{tabular}                                                                                                                                           \\ \hline
\begin{tabular}[c]{@{}l@{}}CONCEPTUAL\\  SIMILARITY\end{tabular}            & \begin{tabular}[c]{@{}l@{}}identical (5)\\ high similarity (4)\\ similar (3)\\ low similarity (2)\\ dissimilar (1)\\ neutral (0)\end{tabular} & Same as above.                                                                                                                                                                                                                                                                                                                                            \\ \hline
\begin{tabular}[c]{@{}l@{}}LEVEL OF \\ ATTENTION\end{tabular}               & \begin{tabular}[c]{@{}l@{}}identical (5)\\ high similarity (4)\\ similar (3)\\ low similarity (2)\\ dissimilar (1)\end{tabular}               & \begin{tabular}[c]{@{}l@{}}The scale is based on words used to describe attentiveness of the relevant \\ public in the relevant case law. The scale is numerical because in some \\ cases, analysts will need to input the average between two numbers.\end{tabular}                                                                                  \\ \hline
CONFUSION                                                                   & \begin{tabular}[c]{@{}l@{}}confusion\\ no confusion\end{tabular}                                                                              & \begin{tabular}[c]{@{}l@{}}This is the final assessment, likely under the heading “global assessment”\\ which provides for whether the trademark sought can be confused with the \\ earlier mark.\end{tabular}                                                                                                                                          \\ \hline
\end{tabular}

\caption{Annotation guideline created by legal professionals to assist lawyers in maintaining consistency for extraction and labelling of relevant sentences from case law.}
\label{annotate:guidelines}
\end{table*}

\begin{table*}[htpb]
\centering
{\begin{tabular}{|l|l|l|}
\hline
\textbf{Annotator \#} & \textbf{Confusion Rules}                                                                                                        & \textbf{No Confusion Rules}                                                                                                                             \\ \hline
\multirow{3}{*}{1}    & $\left(X_{1} \geq 4\right) \wedge\left(\left(X_{2} \geq 4\right) \vee\left(X_{3} \geq 4\right)\right)$                                                    & \multirow{3}{*}{\begin{tabular}[c]{@{}l@{}} $\begin{array}{l}
\left(X_{1} \geq 4\right) \wedge\left(\left(X_{2}<2\right)\right. \\
\left.\vee\left(X_{3}<2\right)\right)
\end{array}$ \end{tabular}}                            \\ \cline{2-2}
                      & $\left(X_{1} \geq 4\right) \vee\left(X_{2} \geq 4\right) \vee\left(X_{3} \geq 4\right) \vee\left(X_{4} \geq 4\right) \vee\left(X_{5} \geq 4\right)$ &                                                                                                                                                         \\ \cline{2-2}
                      & $\left(X_{1}<2\right) \vee\left(X_{2}<2\right) \vee\left(X_{3}<2\right) \vee\left(X_{4}<2\right) \vee\left(X_{5}<2\right)$                              &                                                                                                                                                         \\ \hline
\multirow{2}{*}{2}    & $\left(X_{1}>3\right) \wedge\left(X_{5}>3\right) \wedge\left(X_{2} \geq 3\right) \wedge\left(X_{3} \geq 3\right) \wedge\left(X_{4} \geq 3\right)$      & \multirow{2}{*}{\begin{tabular}[c]{@{}l@{}}$\begin{array}{l}
\left(X_{1} \geq 4\right) \wedge\left(X_{2} \leq 2\right) \wedge \\
\left(X_{3} \leq 2\right) \wedge\left(X_{4} \leq 2\right)
\end{array}$\end{tabular}} \\ \cline{2-2}
                      & $\left(X_{1} \geq 4\right) \wedge\left(X_{2} \leq 2\right) \wedge\left(X_{3} \leq 2\right) \wedge\left(X_{4} \leq 2\right)$                                   &                                                                                                                                                         \\ \hline
\end{tabular}}

\caption{Control group study (assessment of cases by lawyers to establish rules). ${X_{i} \in [1, 5]}$ correspond to ``Similarity of Goods \& Services", ``Visual Similarity", ``Phonetic Similarity", ``Conceptual Similarity", and ``Level of Attention" respectively.}
\label{rules:guidelines}
\end{table*}



\end{document}